\documentclass[a4paper]{article}
\usepackage[T1]{fontenc}
\usepackage{epsfig}
\usepackage{colacl}
\usepackage{verbatim}
\usepackage{amssymb}     
\usepackage{subfigure}
\usepackage{bbm}
\newcounter{parse}
\renewcommand{\theparse}{\arabic{parse}}

\newcounter{algorithm}
\renewcommand{\thealgorithm}{\arabic{algorithm}}

\title{Coaxing Confidences from an Old Friend:\\
Probabilistic Classifications from Transformation Rule Lists
} 

\author{
  
  \begin{tabular}{c@{\hspace*{1cm}}c@{\hspace*{1cm}}c}
    \bf{Radu Florian}\footnotemark
    &
    \bf{John C. Henderson}\footnotemark
    &
    \bf{Grace Ngai}\footnotemark[1] \\
  \end{tabular}\\
  \\
  
  \begin{tabular}{cc}
    \begin{tabular}[t]{c}
      \footnotemark[1] Department of Computer Science\\
      Johns Hopkins University\\
      Baltimore, MD 21218, USA\\
      \{rflorian,gyn\}@cs.jhu.edu
    \end{tabular} &
    \begin{tabular}[t]{c}
      \footnotemark[2] The MITRE Corporation\\
      202 Burlington Road\\
      Bedford, MA 01730, USA\\
      jhndrsn@mitre.org
    \end{tabular} \\
  \end{tabular}
  \setcounter{footnote}{0}
  }

\begin{document}

\newcommand{\calE}{\mathcal{E} }

\maketitle

\begin{abstract}
  Transformation-based learning has been successfully employed to
  solve many natural language processing problems.  It has many
  positive features, but one drawback is that it does not provide
  estimates of class membership probabilities.
  
  In this paper, we present a novel method for obtaining class
  membership probabilities from a transformation-based rule list
  classifier.  Three experiments are presented which measure the
  modeling accuracy and cross-entropy of the probabilistic classifier on
  unseen data and the degree to which the output probabilities from
  the classifier can be used to estimate confidences in its
  classification decisions.
  
  The results of these experiments show that, for the task of text
  chunking\footnote{All the experiments are performed on text
    chunking.  The technique presented is general-purpose, however,
    and can be applied to many tasks for which transformation-based
    learning performs well, without changing the internals of the
    learner.}, the estimates produced by this technique are more
  informative than those generated by a state-of-the-art decision
  tree.

\end{abstract}

\section{Introduction}

In natural language processing, a great amount of work has gone into
the development of machine learning algorithms which extract useful
linguistic information from resources such as dictionaries, newswire
feeds, manually annotated corpora and web pages.  Most of the
effective methods can be roughly divided into rule-based and
probabilistic algorithms.  In general, the rule-based methods have the
advantage of capturing the necessary information in a small and
concise set of rules.  In part-of-speech tagging, for example,
rule-based and probabilistic methods achieve comparable accuracies,
but rule-based methods capture the knowledge in a hundred or so simple
rules, while the probabilistic methods have a very high-dimensional
parameter space (millions of parameters).


One of the main advantages of probabilistic methods, on the other
hand, is that they include a measure of uncertainty in their output.
This can take the form of a probability distribution over potential
outputs, or it may be a ranked list of candidate outputs.  These
uncertainty measures are useful in situations where both the
classification of an sample and the system's confidence in that
classification are needed.  An example of this is a situation in an
ensemble system where ensemble members disagree and a decision must
be made about how to resolve the disagreement.  A similar situation
arises in pipeline systems, such as a system which performs parsing on
the output of a probabilistic part-of-speech tagging.

Transformation-based learning (TBL) \cite{brill95:tagging} is a
successful rule-based machine learning algorithm in natural language
processing.  It has been applied to a wide variety of tasks, including
part of speech tagging \cite{roche94,brill95:tagging},
noun phrase chunking \cite{ramshaw99:basenp}, parsing
\cite{brill96:transformation_parsing,vilain96:parsing}, spelling
correction \cite{mangu97:cssc}, prepositional phrase attachment
\cite{brill94:pp-attach}, dialog act tagging
\cite{samuel98:dialogact}, segmentation and message understanding
\cite{day97:alembic}, often achieving state-of-the-art performance
with a small and easily-understandable list of rules.

In this paper, we describe a novel method which enables a
transformation-based classifier to generate a probability distribution
on the class labels.  Application of the method allows the
transformation rule list to retain the robustness of the
transformation-based algorithms, while benefitting from the advantages
of a probabilistic classifier. The usefulness of the resulting
probabilities is demonstrated by comparison with another
state-of-the-art classifier, the C4.5 decision tree
\cite{quinlan93:c45}.  The performance of our algorithm compares
favorably across many dimensions: it obtains better perplexity and
cross-entropy; an active learning algorithm using our system
outperforms a similar algorithm using decision trees; and finally, our
algorithm has better rejection curves than a similar decision tree.
Section 2 presents the transformation based learning paradigm; Section
3 describes the algorithm for construction of the decision tree
associated with the transformation based list; Section 4 describes the
experiments in detail and Section 5 concludes the paper and outlines
the future work.

\section{Transformation rule lists}


The central idea of transformation-based learning is to learn an
ordered list of rules which progressively improve upon the current
state of the training set. An initial assignment is made based on
simple statistics, and then rules are greedily learned to correct the
mistakes, until no net improvement can be made.

These definitions and notation will be used throughout the paper:
\begin{itemize}
\item $\mathcal{X}$ denotes the {\it sample space};
\item $\mathcal{C}$ denotes the set of possible {\it classifications}
of the samples;
\item The {\it state space} is defined as $\mathcal{S} = \mathcal{X}
\times \mathcal{C}$. 
\item $\pi$ will usually denote a {\it predicate} defined on
$\mathcal{X}$; 
\item A {\it rule} $r$ is defined as a predicate -- class label -- time
tuple, $(\pi, c, t)$, $c \in \mathcal{C}, t \in \mathbbm{N}$, where
$t$ is the learning iteration in which when the rule was learned, its
position in the list.
\item A rule $r = (\pi, c, t)$ {\it applies} to a state
$(x,y)$ if $\pi(x) = true$ and $c \neq y$.
\end{itemize}

\noindent Using a TBL framework to solve a problem assumes the
existence of: 
\begin{itemize}
\item An initial class assignment (mapping from $\mathcal{X}$ to
$\mathcal{S}$).  This can be as simple as the most common class label
in the training set, or it can be the output from another classifier.
\item A set of allowable templates for rules.  These templates
determine the predicates the rules will test, and they have the
biggest influence over the behavior of the system.
\item An objective function for learning.  Unlike in many other learning
algorithms, the objective function for TBL will typically optimize the
evaluation function.  An often-used method is the difference in
performance resulting from applying the rule.
\end{itemize}

At the beginning of the learning phase, the training set is first given
an initial class assignment.  The system then iteratively executes the
following steps:

\begin{enumerate}
\item \label{tblstart} Generate all productive rules.
\item For each rule:
\begin{enumerate}
\item Apply to a copy of the most recent state of the training set.
\item Score the result using the objective function.
\end{enumerate}
\item Select the rule with the best score.
\item Apply the rule to the current state of the training set, updating
it to reflect this change.
\item Stop if the score is smaller than some pre-set threshold $T$.
\item Repeat from Step \ref{tblstart}.
\end{enumerate}

The system thus learns a list of rules in a greedy fashion, according
to the objective function.  When no rule that improves the current
state of the training set beyond the pre-set threshold can be found,
the training phase ends.  During the evaluation phase, the evaluation
set is initialized with the same initial class assignment.  Each rule
is then applied, in the order it was learned, to the evaluation set.
The final classification is the one attained when all rules have been
applied.

\section{Probability estimation with transformation rule lists}

Rule lists are infamous for making {\it hard decisions}, decisions
which adhere entirely to one possibility, excluding all others.  These
hard decisions are often accurate and outperform other types of
classifiers in terms of exact-match accuracy, but because they do not
have an associated probability, they give no hint as to when they
might fail.  In contrast, probabilistic systems make {\it soft
  decisions} by assigning a probability distribution over all possible
classes.

There are many applications where soft decisions prove useful.  In
situations such as \textit{active learning}, where a small number of
samples are selected for annotation, the probabilities can be used to
determine which examples the classifier was most unsure of, and hence
should provide the most extra information.  A probabilistic system can
also act as a filter for a more expensive system or a human expert
when it is permitted to \textit{reject} samples.  Soft decision-making
is also useful when the system is one of the components in a larger
decision-making process, as is the case in speech recognition systems
\cite{bahl89:treelm}, or in an ensemble system like AdaBoost
\cite{freund97:adaboost}.  There are many other applications in which
a probabilistic classifier is necessary, and a non-probabilistic
classifier cannot be used instead.


\subsection{Estimation via conversion to decision tree}

The method we propose to obtain probabilistic classifications from a
transformation rule list involves dividing the samples into
equivalence classes and computing distributions over each equivalence
class.  At any given point in time $i$, each sample $x$ in the
training set has an associated state $s_i(x) = (x,y)$.  Let $R(x)$ to
be the set of rules $r_i$ that applies to the state $s_i(x)$,
\[ R(x) = \{ r_i \in \mathcal{R} | r_i \mbox{ applies to } s_i(x) \} \]
An equivalence class consists of all the samples $x$ that have the
same $R(x)$.  Class probability assignments are then estimated using
statistics computed on the equivalence classes.

\begin{figure}[htbp]
\centering
  \epsfig{file=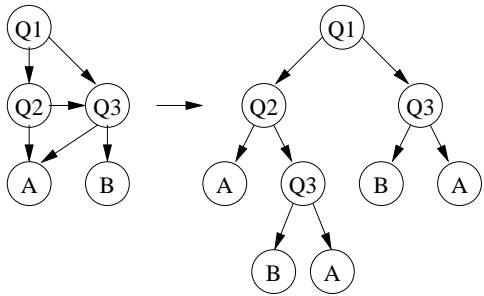,width= .4  \textwidth}

\caption{\label{fig:rltodt}  Converting the transformation rule list
from Table \ref{table:samplerl} to a decision tree.}
\end{figure}

An illustration of the conversion from a rule list to a decision tree
is shown below.  Table \ref{table:samplerl} shows an example
transformation rule list. It is straightforward to convert this rule
list into a decision pylon \cite{bahl89:treelm}, which can be used to
represent all the possible sequences of labels assigned to a sample
during the application of the TBL algorithm. The decision pylon associated with
this particular rule list is displayed on the left side of Figure
\ref{fig:rltodt}. The decision tree shown on the right side of Figure
\ref{fig:rltodt} is constructed such that the samples stored in any
leaf have the same class label sequence as in the displayed decision pylon.  In
the decision pylon, ``no'' answers go straight down; in the decision tree,
``yes'' answers take the right branch.  Note that a one rule in the
transformation rule list can often correspond to more than one node in
the decision tree.

\begin{table}[htbp]
{\centering \begin{tabular}{|c|}
\hline
Initial label = A\\
\hline
{\bf If} Q1 {\bf and} label=A {\bf then} label$\leftarrow$B\\
\hline
{\bf If} Q2 {\bf and} label=A {\bf then} label$\leftarrow$B\\
\hline
{\bf If} Q3 {\bf and} label=B {\bf then} label$\leftarrow$A\\
\hline 
\end{tabular}\par}

\caption{\label{table:samplerl}Example of a Transformation Rule List.}
\end{table} 

The conversion from a transformation rule list to a decision tree is
presented as a recursive procedure.  The set of samples in the
training set is transformed to a set of states by applying the initial
class assignments.  A node $n$ is created for each of the initial
class label assignments $c$ and all states labeled $c$ are
assigned to $n$.

The following recursive procedure is invoked with an initial ``root''
node, the complete set of states (from the corpus) and the whole
sequence of rules learned during training:



\smallskip
\vspace*{2mm}
\noindent\textbf{Algorithm: RuleListToDecisionTree \\\hspace*{2cm}(RLTDT)}\\
\textbf{Input:}
\begin{itemize}
\item A set $\mathcal{B}$ of $N$ states $\langle(x_{1},y_{1})\ldots
(x_{N},y_{N})\rangle$ with labels $y_{i} \in C$;
\item A set $\mathcal{R}$ of $M$ rules $\langle r_{0},r_{1}\ldots
r_{M}\rangle$ where $r_{i} = (\pi_{i},y_{i},i)$.
\end{itemize}

\noindent\textbf{Do:}
\begin{enumerate}

\item \label{tbldt:start}If $\mathcal{R}$ is empty, the end of the
rule list has been reached. Create a leaf node, $n$, and estimate the
probability class distribution based on the true classifications of
the states in $\mathcal{B}$.  Return $n$.
\item Let $r_j = (\pi_j, y_j, j)$ be the lowest-indexed rule in
$\mathcal{R}$. Remove it from $\mathcal{R}$.
\item Split the data in $\mathcal{B}$ using the predicate $\pi_j$ and
the current hypothesis such that samples on which $\pi_j$ returns
\textit{true} are on the right of the split:
\begin{eqnarray*}
\mathcal{B}_L &=& \left\{ x \in \mathcal{B} | \pi_j(x) = false \right\}\\
\mathcal{B}_R &=& \left\{ x \in \mathcal{B} | \pi_j(x) = true \right\}
\end{eqnarray*}
\item If $|\mathcal{B}_{L}|>K$ and $|\mathcal{B}_{R}|>K$, the split is
acceptable:  
\begin{enumerate}
\item Create a new internal node, $n$;
\item Set the question: $q(n)=\pi_j$;
\item Create the left child of $n$ using a recursive call to
$RLTDT(\mathcal{B}_{L}, \mathcal{R})$;
\item Create the right child of $n$ using a recursive call to
$RLTDT(\mathcal{B}_{R},\mathcal{R})$; 
\item Return node $n$.
\end{enumerate}
Otherwise, no split is performed using $r_j$.  Repeat from Step
\ref{tbldt:start}.
\end{enumerate}
\noindent 
The parameter $K$ is a constant that determines the minimum weight 
that a leaf is permitted to have, effectively pruning the tree during
construction.  In all the experiments, $K$ was set to 5.

\subsection{Further growth of the decision tree}

When a rule list is converted into a decision tree, there are often
leaves that are inordinately heavy because they contain a large number
of samples.  Examples of such leaves are those containing samples
which were never transformed by any of the rules in the rule list.
These populations exist either because they could not be split up
during the rule list learning without incurring a net penalty, or
because any rule that acts on them has an objective function score of
less than the threshold $T$.  This is sub-optimal for estimation
because when a large portion of the corpus falls into the same
equivalence class, the distribution assigned to it reflects only the
mean of those samples.  The undesirable consequence is that all of
those samples are given the same probability distribution.

To ameliorate this problem, those samples are partitioned into smaller
equivalence classes by further growing the decision tree.  Since a
decision tree does not place all the samples with the same current
label into a single equivalence class, it does not get stuck in the
same situation as a rule list \,---\, in which no change in the current
state of corpus can be made without incurring a net loss in
performance.

Continuing to grow the decision tree that was converted from a rule
list can be viewed from another angle.  A highly accurate prefix tree
for the final decision tree is created by tying questions together
during the first phase of the growth process (TBL).  Unlike
traditional decision trees which select splitting questions for a node
by looking only at the samples contained in the local node, this
decision tree selects questions by looking at samples contained in all
nodes on the frontier whose paths have a suffix in common.  An
illustration of this phenomenon can be seen in Figure
\ref{fig:rltodt}, where the choice to split on Question 3 was made
from samples which tested false on the predicate of Question 1,
together with samples which tested false on the predicate of Question
2.  The result of this is that questions are chosen based on a much
larger population than in standard decision tree growth, and therefore
have a much greater chance of being useful and generalizable.  This
alleviates the problem of over-partitioning of data, which is a
widely-recognized concern during decision tree growth.

The decision tree obtained from this conversion can be grown further.
When the rule list $\mathcal{R}$ is exhausted at Step
\ref{tbldt:start}, instead of creating a leaf node, continue splitting
the samples contained in the node with a decision tree induction
algorithm.  The splitting criterion used in the experiments is the
information gain measure.


\section{Experiments}

Three experiments that demonstrate the effectiveness and
appropriateness of our probability estimates are presented in this
section.  The experiments are performed on text chunking, a subproblem
of syntactic parsing.  Unlike full parsing, the sentences are divided
into non-overlapping phrases, where each word belongs to the lowest
parse constituent that dominates it.

The data used in all of these experiments is the CoNLL-2000 phrase
chunking corpus \cite{conll-website}.  The corpus consists of sections
15-18 and section 20 of the Penn Treebank \cite{penntreebank}, and is
pre-divided into a 8936-sentence (211727 tokens) training set and a
2012-sentence (47377 tokens) test set.  The chunk tags are derived
from the parse tree constituents, and the part-of-speech tags were
generated by the Brill tagger \cite{brill95:tagging}.  

As was noted by Ramshaw \& Marcus \shortcite{ramshaw99:basenp}, text
chunking can be mapped to a tagging task, where each word is tagged
with a chunk tag representing the phrase that it belongs to.  An
example sentence from the corpus is shown in Table \ref{table:example}.
\begin{table}
\begin{center}
\begin{tabular}{|c|c|c|}
\hline
Word & POS tag & Chunk Tag\\
\hline
A.P. &  NNP & B-NP\\
Green & NNP & I-NP\\
currently & RB & B-ADVP\\
has & VBZ & B-VP\\
2,664,098 & CD & B-NP\\
shares & NNS & I-NP\\
outstanding & JJ & B-ADJP\\
. & . & O\\
\hline
\end{tabular}\\
\label{table:example}
\caption{Example of a sentence with chunk tags}
\end{center}
\end{table}
As a contrasting system, our results are compared with those produced
by a C4.5 decision tree system (henceforth C4.5).  The reason for
using C4.5 is twofold: firstly, it is a widely-used algorithm which
achieves state-of-the-art performance on a broad variety of tasks; and
secondly, it belongs to the same class of classifiers as our converted
transformation-based rule list (henceforth TBLDT).

To perform a fair evaluation, extra care was taken to ensure that both
C4.5 and TBLDT explore as similar a sample space as possible.  The
systems were allowed to consult the word, the part-of-speech, and the
chunk tag of all examples within a window of 5 positions (2 words on
either side) of each target example.\footnote{The TBL templates are
similar to those used in \newcite{ramshaw99:basenp}.}  Since multiple
features covering the entire vocabulary of the training set would be
too large a space for C4.5 to deal with, in all of experiments where
TBLDT is directly compared with C4.5, the word types that both systems
can include in their predicates are restricted to the most
``ambiguous'' 100 words in the training set, as measured by the number
of chunk tag types that are assigned to them.  The initial prediction
was made for both systems using a class assignment based solely on the
part-of-speech tag of the word.

Considering chunk tags within a contextual window of the target word
raises a problem with C4.5.  A decision tree generally trains on
independent samples and does not take into account changes of any
features in the context.  In our case, the samples are dependent; the
classification of sample $i$ is a feature for sample $i+1$, which
means that changing the classification for sample $i$ affects the
context of sample $i+1$.  To address this problem, the C4.5 systems
are trained with the correct chunks in the left context.  When the
system is used for classification, input is processed in a
left-to-right manner; and the output of the system is fed forward to
be used as features in the left context of following samples.  Since
C4.5 generates probabilities for each classification decision, they
can be redirected into the input for the next position.  Providing the
decision tree with this confidence information effectively allows it
to perform a limited search over the entire sentence.

C4.5 does have one advantage over TBLDT, however.  A decision tree can
be trained using the {\em subsetting} feature, where questions asked
are of the form: ``does feature $f$ belong to the set $F$?''.  This is
not something that a TBL can do readily, but since the objective is
in comparing TBLDT to another state-of-the-art system, this feature
was enabled.

\subsection{Evaluation Measures}
The most commonly used measure for evaluating tagging tasks is tag accuracy. It is defined as 
\[
\mbox{Accuracy} = \frac {\mbox{\# of correctly tagged examples}}{\mbox{\# of examples}}
\]
In syntactic parsing, though, since the task is to identify the
phrasal components, it is more appropriate to measure the precision
and recall:
\begin{eqnarray*}
\mbox{Precision} &=& \frac{\mbox{\# of correct
proposed phrases}}{\mbox{\# of proposed phrases}}\\
\mbox{Recall} &=& \frac{\mbox{\# of correct proposed
phrases}}{\mbox{\# of correct phrases}}
\end{eqnarray*}
To facilitate the comparison of systems with different precision and
recall, the F-measure metric is computed as a weighted harmonic mean
of precision and recall:
\begin{eqnarray*}
F_\beta &=&
\frac{(\beta^2+1)\times\mbox{Precision}\times\mbox{Recall}}{\beta^2\times\mbox{Precision}+\mbox{Recall}}\\
\end{eqnarray*}
The $\beta$ parameter is used to give more weight to precision or
recall, as the task at hand requires.  In all our experiments, $\beta$
is set to 1, giving equal weight to precision and recall.

The reported performances are all measured with the evaluation tool
provided with the CoNLL corpus \cite{conll-website}.


\begin{figure*}
\subfigure[F-measure vs. number of words in training set]{\resizebox*{0.47\textwidth}{5.2cm}{\rotatebox{-90}{\includegraphics{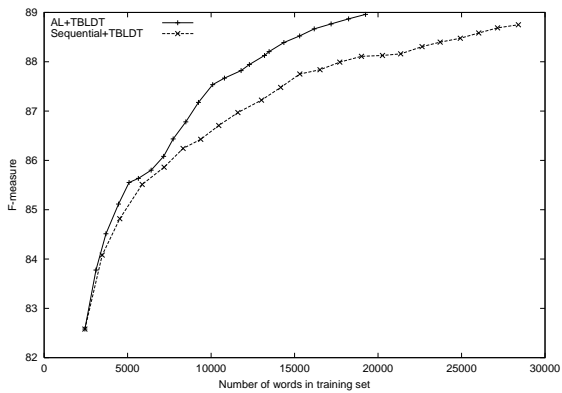}}}}
\subfigure[Chunk Accuracy vs. number of words in training set] {\resizebox*{0.47\textwidth}{5.2cm}{\rotatebox{-90}{\includegraphics{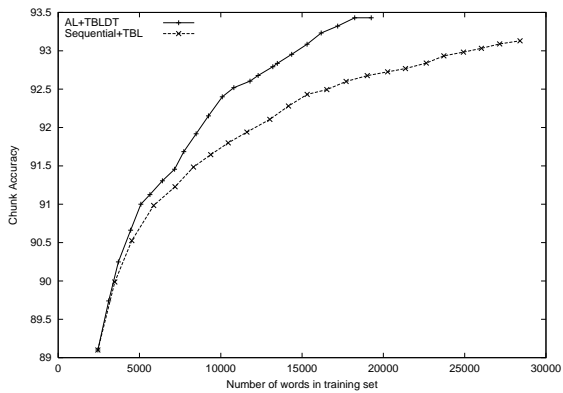}}}}

\caption{Performance of the TBLDT system versus sequential choice.\label{TBL-sequential}}
\end{figure*}

\subsection{Active Learning}
To demonstrate the usefulness of obtaining probabilities from a
transformation rule list, this section describes an application which
utilizes these probabilities, and compare the resulting performance of
the system with that achieved by C4.5.

Natural language processing has traditionally required large amounts
of annotated data from which to extract linguistic properties.
However, not all data is created equal: a normal distribution of
annotated data contains much redundant
information. \newcite{seung92:query_by_committee} and
\newcite{freund97:query_by_committee} proposed a theoretical active
learning approach, where samples are intelligently selected for
annotation.  By eliminating redundant information, the same
performance can be achieved while using fewer resources.  Empirically,
active learning has been applied to various NLP tasks such as text
categorization
\cite{lewis94:active_learning,lewis94:heterogeneous,liere97:active_learning_textcat},
part-of-speech tagging
\cite{dagan95:active_learning,engelson96:active_learning_POS}, and
base noun phrase chunking \cite{ngai00:al_basenp}, resulting in
significantly large reductions in the quantity of data needed to
achieve comparable performance.

This section presents two experimental results which show the
effectiveness of the probabilities generated by the TBLDT.  The first
experiment compares the performance achieved by the active learning
algorithm using TBLDT with the performance obtained by selecting
samples sequentially from the training set.  The second experiment
compares the performances achieved by TBLDT and C4.5 training on
samples selected by active learning.

The following describes the active learning algorithm used in the
experiments:
\begin{enumerate}
\item Label an initial $T_1$ sentences of the corpus;
\item \label{al:repeat}Use the machine learning algorithm (C4.5 or
TBLDT) to obtain chunk probabilities on the rest of the training data;
\item Choose $T_2$ samples from the rest of the training
set, specifically the samples that optimize an evaluation function
$f$, based on the class distribution probability of each sample; 
\item Add the samples, including their ``true''
  classification\footnote{The true (reference or gold standard)
    classification is available in this experiment. In an annotation
    situation, the samples are sent to human annotators for labeling.}
  to the training pool and retrain the system;
\item If a desired number of samples is reached, stop, otherwise
repeat from Step \ref{al:repeat}.
\end{enumerate}
The evaluation function $f$ that was used in our experiments is:
\[
f\left( S\right) = \frac{1}{|S|}\sum_{i=1}^{|S|} H\left(C|S,i\right)\]
where \( H\left( C|S,i\right)  \) is the entropy of the chunk
probability distribution associated with the word index \(i\) in
sentence \(S\). 
\begin{figure*}
\subfigure[F-measure vs. number of words in training set]{\resizebox*{0.47\textwidth}{!}{\rotatebox{-90}{\includegraphics{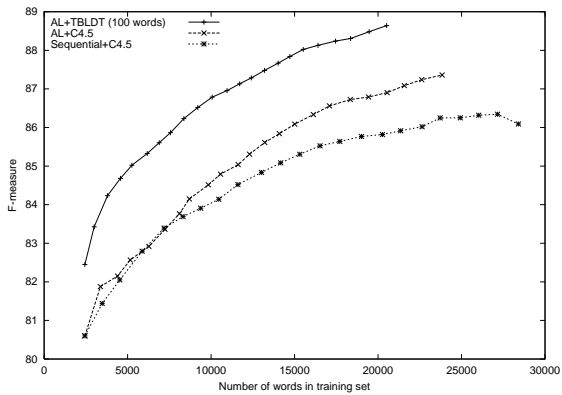}}}}
\subfigure[Accuracy vs. number of words in training set]{\resizebox*{0.47\textwidth}{!}{\rotatebox{-90}{\includegraphics{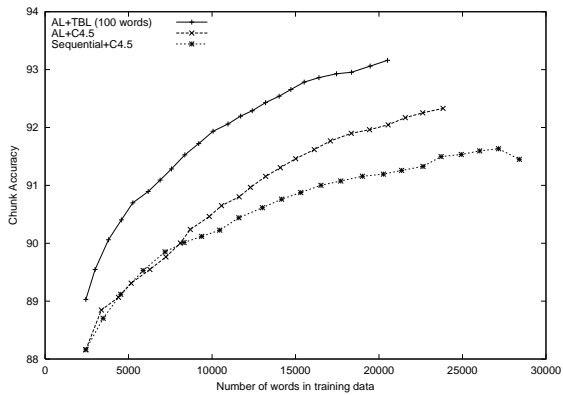}}}}

\caption{Performance of the TBLDT system versus the DT system\label{TBL-DT-comp}}
\end{figure*}

Figure \ref{TBL-sequential} displays the performance (F-measure and
chunk accuracy) of a TBLDT system trained on samples selected by
active learning and the same system trained on samples selected
sequentially from the corpus versus the number of words in the
annotated training set. At each step of the iteration, the active
learning-trained TBLDT system achieves a higher accuracy/F-measure,
or, conversely, is able to obtain the same performance level with less
training data.  Overall, our system can yield the same performance as
the sequential system with 45\% less data, a significant reduction in
the annotation effort.

Figure \ref{TBL-DT-comp} shows a comparison between two active
learning experiments: one using TBLDT and the other using
C4.5.\footnote{As mentioned earlier, both the TBLDT and C4.5 were
limited to the same 100 most ambiguous words in the corpus to ensure
comparability.}  For completeness, a sequential run using C4.5 is also
presented. Even though C4.5 examines a larger space than TBLDT by
utilizing the feature subset predicates, TBLDT still performs
better. The difference in accuracy at 26200 words (at the end of the
active learning run for TBLDT) is statistically significant at a
0.0003 level.

As a final remark on this experiment, note that at an annotation level
of 19000 words, the fully lexicalized TBLDT outperformed the C4.5
system by making 15\% fewer errors.


\subsection{Rejection curves} \label{rejection_curves}
\begin{figure*}[htbp]
\subfigure[Subcorpus (batch) rejection \label{fig:batch_rejection}] {
    \epsfig{file= 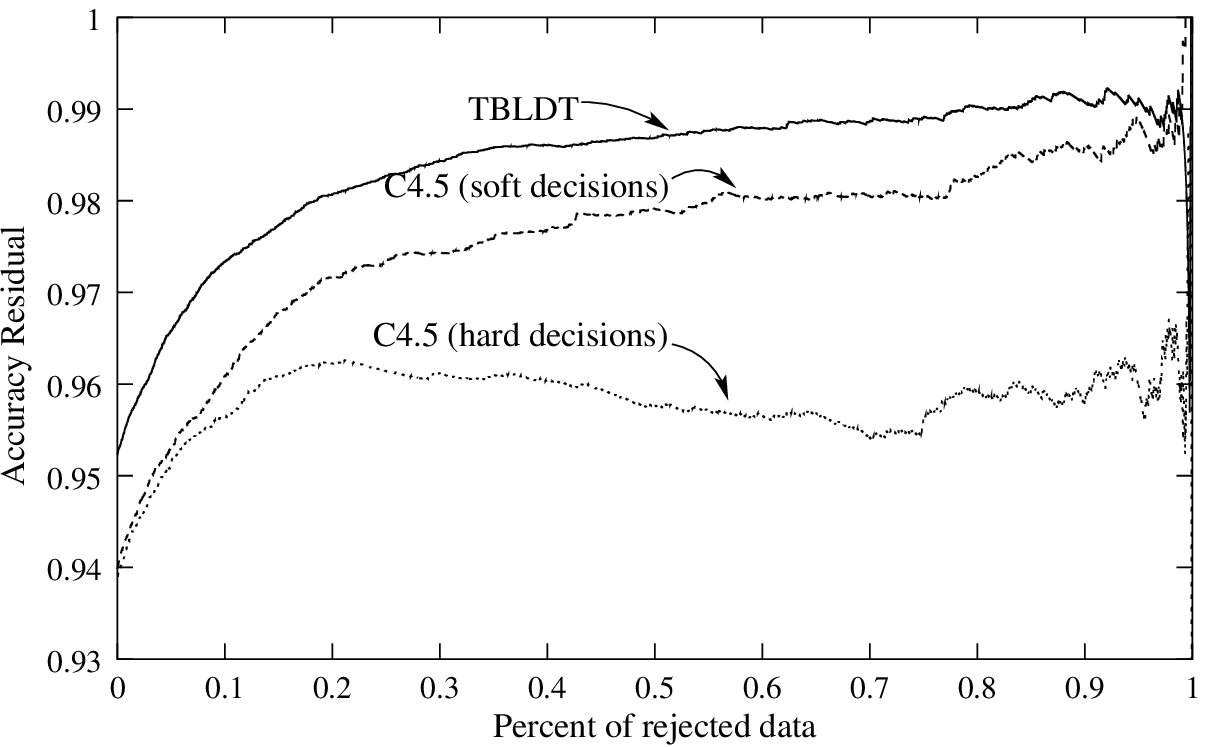, width=.45 \textwidth, height=5.2cm}
}
\subfigure[Threshold (online) rejection \label{fig:online_rejection}] {
\epsfig{file=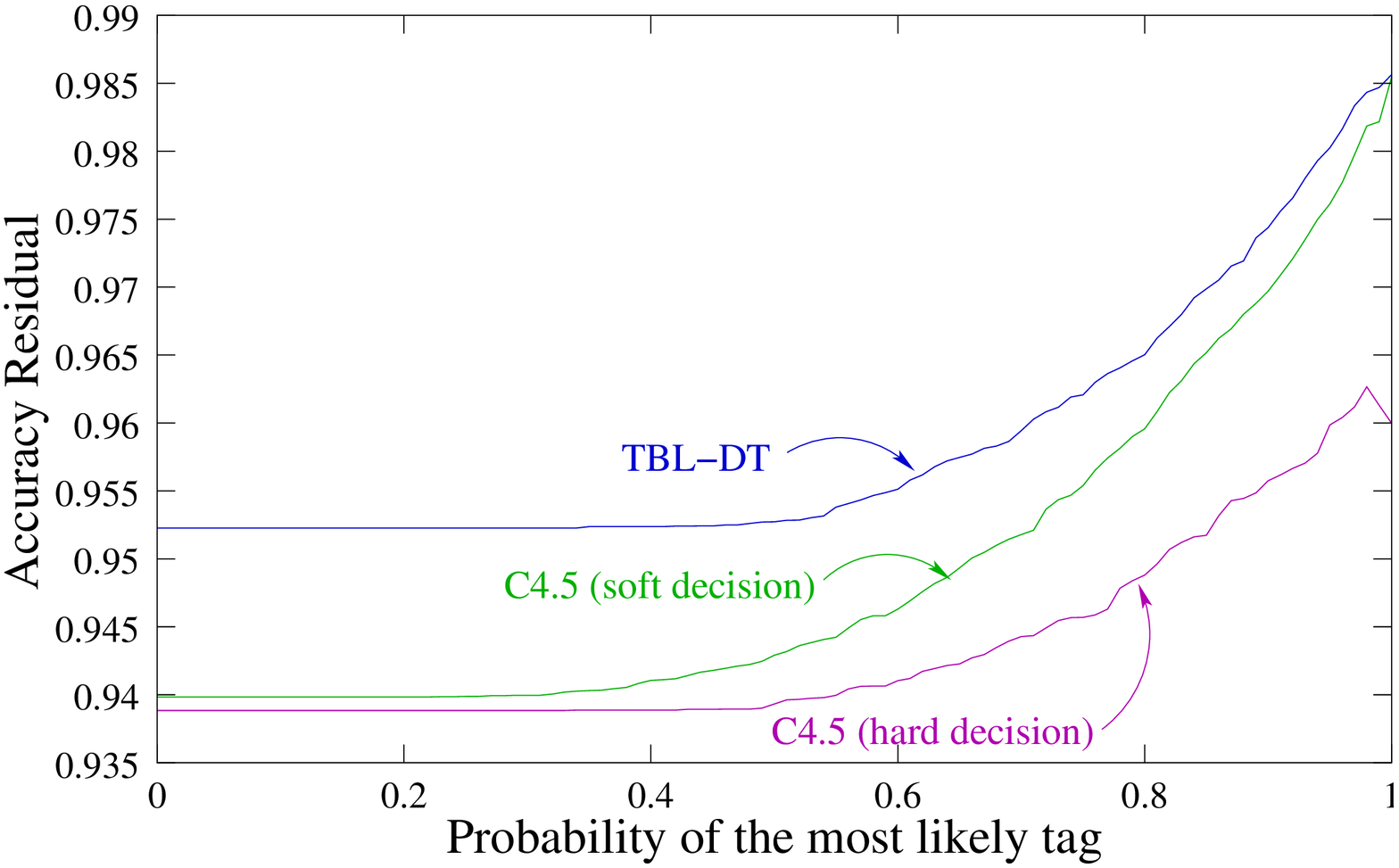,width= .45  \textwidth, height=5.2cm}
}
\caption{Rejection curves.}
\label{fig:rejection}
\end{figure*}

It is often very useful for a classifier to be able to offer
confidence scores associated with its decisions. Confidence scores are
associated with the probability \(P( C(x) \textnormal{ correct}|x)\)
where \(C(x) \) is the classification of sample \(x\). These scores
can be used in real-life problems to \textit{reject} samples that the
the classifier is not sure about, in which case a better observation,
or a human decision, might be requested.  The performance
of the classifier is then evaluated on the samples that were not
rejected.  This experiment framework is well-established in machine
learning and optimization research
\cite{multiclassecoc,priebe99:_optim_mine_class_perfor}.

Since non-probabilistic classifiers do not offer any insights into how
sure they are about a particular classification, it is not easy to
obtain confidence scores from them.  A probabilistic classifier, in
contrast, offers information about the class probability distribution
of a given sample. Two measures that can be used in generating
confidence scores are proposed in this section.

The first measure, the entropy $H$ of the class probability
distribution of a sample \(x\), \( C(x)=\{p(c_{1}|x),p(c_{2}|x)\ldots
p(c_{k}|x)\} \), is a measure of the uncertainty in the distribution:

\[
H(C(x))=-\sum _{i=1}^{k}p(c_{i}|x)\log _{2}p(c_{i}|x)
\]

The higher the entropy of the distribution of class probability
estimates, the more uncertain the classifier is of its classification.
The samples selected for rejection are chosen by sorting the data
using the entropies of the estimated probabilities, and then selecting
the ones with highest entropies.  The resulting curve is a measure of
the correlation between the true probability distribution and the one
given by the classifier.

Figure \ref{fig:batch_rejection} shows the rejection curves for the
TBLDT system and two C4.5 decision trees -- one which receives a
probability distribution as input (``soft'' decisions on the left
context) , and one which receives classifications (``hard'' decisions
on all fields). At the left of the curve, no samples are rejected; at
the right side, only the samples about which the classifiers were most
certain are kept (the samples with minimum entropy). Note that the
y-values on the right side of the curve are based on less data,
effectively introducing wider variance in the curve as it moves right.

As shown in Figure \ref{fig:batch_rejection}, the C4.5 classifier that
has access to the left context chunk tag probability distributions
behaves better than the other C4.5 system, because this information
about the surrounding context allows it to effectively perform a
shallow search of the classification space. The TBLDT system, which
also receives a probability distribution on the chunk tags in the left
context, clearly outperforms both C4.5 systems at all rejection levels.


The second proposed measure is based on the probability of the most
likely tag. The assumption here is that this probability is
representative of how certain the system is about the classification.
The samples are put in bins based on the probability of the most
likely chunk tag, and accuracies are computed for each bin (these bins
are cumulative, meaning that a sample will be included in all the bins
that have a lower threshold than the probability of its most likely
chunk tag).  At each accuracy level, a sample will be rejected if the
probability of its most likely chunk is below the accuracy level. The
resulting curve is a measure of the correlation between the true
distribution probability and the probability of the most likely chunk
tag, i.e. how appropriate those probabilities are as confidence
measures. Unlike the first measure mentioned before, a threshold
obtained using this measure can be used in an online manner to
identify the samples of whose classification the system is confident.

Figure \ref{fig:online_rejection} displays the rejection curve for the
second measure and the same three systems.  TBLDT again outperforms
both C4.5 systems, at all levels of confidence.

In summary, the TBLDT system outperforms both C4.5 systems presented,
resulting in fewer rejections for the same performance, or,
conversely, better performance at the same rejection rate.


\subsection{Perplexity and Cross Entropy}

Cross entropy is a goodness measure for probability estimates that
takes into account the accuracy of the estimates as well as the
classification accuracy of the system. It measures the performance of
a system trained on a set of samples distributed according to the
probability distribution $p$ when tested on a set following a
probability distribution $q$. More specifically, we utilize
conditional cross entropy, which is defined as

\begin{table}
\centering
 \begin{tabular}{|c|c|c|}
\hline 
Model&
Perplexity&
Cross Entropy\\
\hline 
\hline 
TBLDT&
1.2944&
0.2580\\
\hline 
DT+probs&
1.4150&
0.3471\\
\hline 
DT&
1.4568&
0.3763\\
\hline 
\end{tabular}
\caption{Cross entropy and perplexities for two C4.5 systems and the TBLDT system}
\label{table:perplex}
\end{table}

\[
H\left(C|X\right) = -\sum_{x \in \mathcal{X}} q(x) \cdot \sum_{c \in \mathcal{C}} q(c|x) \cdot \log_2 p(c|x)
\]
where $\mathcal{X}$ is the set of examples and $\mathcal{C}$ is the set of chunk tags, 
$q$ is the probability distribution on the test document and $p$ is the probability distribution
on the train corpus. 


The cross entropy metric fails if any outcome is given zero
probability by the estimator.  To avoid this problem, estimators are
``smoothed'', ensuring that novel events receive non-zero
probabilities.  A very simple smoothing technique (interpolation with
a constant) was used for all of these systems. 

A closely related measure is \textit{perplexity}, defined as
\[
P = 2^{ H(C|X) }
\]
The cross entropy and perplexity results for the various estimation
schemes are presented in Table \ref{table:perplex}. The TBLDT
outperforms both C4.5 systems, obtaining better cross-entropy and
chunk tag perplexity. This shows that the overall probability
distribution obtained from the TBLDT system better matches the true
probability distribution. This strongly suggests that probabilities
generated this way can be used successfully in system combination
techniques such as voting or boosting.




\subsection{Chunking performance}

It is worth noting that the transformation-based system used in the
comparative graphs in Figure \ref{TBL-DT-comp} was not running at full
potential.  As described earlier, the TBLDT system was only allowed to
consider words that C4.5 had access to.  However, a comparison between
the corresponding TBLDT curves in Figures \ref{TBL-sequential} (where
the system is given access to all the words) and \ref{TBL-DT-comp}
show that a transformation-based system given access to all the words
performs better than the one with a restricted lexicon, which in turn
outperforms the best C4.5 decision tree system both in terms of
accuracy and F-measure.


\begin{table}
\small
\centering
\begin{tabular}{|@{\hspace{2pt}}c@{\hspace{2pt}}|@{\hspace{2pt}}c@{\hspace{2pt}}|@{\hspace{2pt}}c@{\hspace{2pt}}|@{\hspace{2pt}}c@{\hspace{2pt}}|@{\hspace{2pt}}c@{\hspace{2pt}}|}
\hline 
\textbf{\parbox{15mm}{\centering Chunk \\Type}}&
\textbf{\parbox{15mm}{\centering Accuracy \\(\%)}}&
\textbf{\parbox{15mm}{\centering Precision \\(\%)}}&
\textbf{\parbox{15mm}{\centering Recall\\ (\%)}}&
\textbf{F$_1$}\\
\hline 
\hline 
Overall &  95.23 &  92.02 & 92.50 &  92.26 \\
\hline
             ADJP & - & 75.69&  68.95&  72.16 \\
\hline
             ADVP & - & 80.88&  78.64&  79.74 \\
\hline
            CONJP & - & 40.00&  44.44&  42.11 \\
\hline
             INTJ & - & 50.00&  50.00&  50.00 \\
\hline
              LST & - &  0.00&   0.00&   0.00 \\
\hline
               NP & - & 92.18&  92.72&  92.45 \\
\hline
               PP & - & 95.89&  97.90&  96.88 \\
\hline
              PRT & - & 67.80&  75.47&  71.43 \\
\hline
             SBAR & - & 88.71&  82.24&  85.35 \\
\hline
               VP & - & 92.00&  92.87&  92.44 \\
\hline
\end{tabular}
\caption{Performance of TBLDT on the CoNLL Test Set
\label{table:fullResult}} 
\end{table}


Table \ref{table:fullResult} shows the performance of the TBLDT system
on the full CoNLL test set, broken down by chunk type.  Even though
the TBLDT results could not be compared with other published results
on the same task and data (CoNLL will not take place until September
2000), our system significantly outperforms a similar system trained
with a C4.5 decision tree, shown in Table \ref{table:DT}, both in
chunk accuracy and F-measure.

\begin{table}
\small
\centering
\begin{tabular}{|@{\hspace{2pt}}c@{\hspace{2pt}}|@{\hspace{2pt}}c@{\hspace{2pt}}|@{\hspace{2pt}}c@{\hspace{2pt}}|@{\hspace{2pt}}c@{\hspace{2pt}}|@{\hspace{2pt}}c@{\hspace{2pt}}|}
\hline 
\textbf{\parbox{15mm}{\centering Chunk\\Type}}&
\textbf{\parbox{15mm}{\centering Accuracy \\ (\%)}}&
\textbf{\parbox{1.5cm}{\centering Precision\\ (\%)}}&
\textbf{\parbox{1.5cm}{\centering Recall\\ (\%)}}&
\textbf{F$_1$}\\
\hline 
\hline 
Overall&
93.80&
90.02&
90.26&
90.14\\
\hline 
ADJP&
-&
65.58&
64.38&
64.98\\
\hline 
ADVP&
-&
74.14&
76.79&
75.44\\
\hline 
CONJP&
-&
33.33&
33.33&
33.33\\
\hline 
INTJ&
-&
50.00&
50.00&
50.00\\
\hline 
LST&
-&
0.00&
0.00&
0.00\\
\hline 
NP&
-&
91.00&
90.93&
90.96\\
\hline 
PP&
-&
92.70&
96.36&
94.50\\
\hline 
PRT&
-&
71.13&
65.09&
67.98\\
\hline 
SBAR&
-&
86.35&
61.50&
71.83\\
\hline 
VP&
-&
90.71&
91.22&
90.97\\
\hline 
\end{tabular}
\caption{Performance of C4.5 on the CoNLL Test Set\label{table:DT}}
\end{table}


\section{Conclusions}

In this paper we presented a novel way to convert transformation rule
lists, a common paradigm in natural language processing, into a form
that is equivalent in its classification behavior, but is capable of
providing probability estimates.  Using this approach, favorable
properties of transformation rule lists that makes them popular for
language processing are retained, while the many advantages of a
probabilistic system are gained.

To demonstrate the efficacy of this approach, the resulting
probabilities were tested in three ways: directly measuring the
modeling accuracy on the test set via cross entropy, testing the
goodness of the output probabilities in a active learning algorithm,
and observing the rejection curves attained from these probability
estimates. The experiments clearly demonstrate that the resulting
probabilities perform at least as well as the ones generated by C4.5
decision trees, resulting in better performance in all cases. This
proves that the resulting probabilistic classifier is as least as good
as other state-of-the-art probabilistic models.

The positive results obtained suggest that the probabilistic
classifier obtained from transformation rule lists can be successfully
used in machine learning algorithms that require soft-decision
classifiers, such as boosting or voting.  Future research will include
testing the behavior of the system under AdaBoost
\cite{freund97:adaboost}.  We also intend to investigate the effects
that other decision tree growth and smoothing techniques may have on
continued refinement of the converted rule list.


\section{Acknowledgements}
We thank Eric Brill, Fred Jelinek and David Yarowsky for their
invaluable advice and suggestions.  In addition we would like to thank
David Day, Ben Wellner and the anonymous reviewers for their useful
comments and suggestions on the paper.

The views expressed in this paper are those of the authors and do not
necessarily reflect the views of the MITRE Corporation.  It was
performed as a collaborative effort at both MITRE and the Center for
Language and Speech Processing, Johns Hopkins University, Baltimore,
MD.  It was supported by NSF grants numbered IRI-9502312 and
IRI-9618874, as well as the MITRE-Sponsored Research program.

\bibliographystyle{acl} 
\bibliography{tbldt}

\end{document}